%% file: main.tex
\documentclass{article} 
\usepackage{iclr2023_conference,times}

\iclrfinalcopy

\definecolor{mydarkblue}{rgb}{0,0.08,0.45}
\usepackage[colorlinks, anchorcolor=white, linkcolor=mydarkblue, urlcolor=mydarkblue, citecolor=mydarkblue]{hyperref}
\usepackage{url}
\usepackage{url}

\title{GAR-meets-RAG Paradigm for Zero-Shot\\ Information Retrieval}

\input{0-header}

\author{
\small{Daman Arora\thanks{Equal Contribution.} , Anush Kini$^*$, Sayak Ray Chowdhury, Nagarajan Natarajan, Gaurav Sinha, Amit Sharma}\\
\small{Microsoft Research, India}\\
\small{\texttt{\{t-damanarora,t-anushkini,t-sayakr,nagarajn,gauravsinha,amshar\}@microsoft.com}} \\
}

\begin{document}

\maketitle

\begin{abstract}
Given a query and a document corpus, the information retrieval (IR) task is to output a ranked list of relevant documents. Combining large language models (LLMs) with embedding-based retrieval models, recent work shows promising results on the \textit{zero-shot} retrieval problem, i.e., no access to labeled data from the target domain. 
Two such popular paradigms are generation-augmented retrieval or GAR (generate additional context for the query and then retrieve), and retrieval-augmented generation or RAG (retrieve relevant documents as context and then generate answers). The success of these paradigms hinges on (i) high-recall retrieval models, which are difficult to obtain in the zero-shot setting, and (ii) high-precision (re-)ranking models which typically need a good initialization. In this work, we propose a novel GAR-meets-RAG recurrence formulation that overcomes the challenges of existing paradigms. Our method iteratively improves retrieval (via GAR) and rewrite (via RAG) stages in the zero-shot setting. A key design principle is that the rewrite-retrieval stages improve the recall of the system and a final re-ranking stage improves the precision.

 We conduct extensive experiments on zero-shot passage retrieval benchmarks, BEIR and TREC-DL. Our method
 establishes a new state-of-the-art in the BEIR benchmark, outperforming previous best results in \recall{}@100 and \ndcg{}@10 metrics on 6 out of 8 datasets, with up to 17\% relative gains over the previous best. 
\end{abstract}

\input{1-intro}
\input{3-problem}

\input{4-method}
\input{5.5-table_beir}
\input{5.6-table_trecdl}
\input{5-evalsetup}
\input{5-results}
\input{5-ablation}
\input{2-related}
\input{6-conclusion}

\bibliography{main}
\bibliographystyle{iclr2023_conference}

\appendix
\clearpage
\input{7-appendix}

\end{document}

%% file: 0-header.tex
\usepackage{amsmath}
\usepackage{amssymb}
\usepackage{mathtools}
\usepackage{amsthm}
\usepackage{dsfont}
\usepackage{algorithm}
\usepackage{algpseudocode}

\usepackage{mathrsfs}
\usepackage{mathnotations}
\usepackage{subcaption, wrapfig}

\usepackage{xcolor} 
\usepackage{xspace}
\usepackage{mdframed} 

\definecolor{palegray}{rgb}{0.9, 0.9, 0.9}

\usepackage{tikz}
\usepackage{tablefootnote}
\usetikzlibrary{fit, positioning, arrows.meta, calc, shadows}

\usepackage{booktabs}

\newcommand*\circled[1]{\tikz[baseline=(char.base)]{
            \node[shape=circle,draw,inner sep=2pt,fill=palegray] (char) {#1};}}

\newcommand{\stc}[1]{{\color[rgb]{0.1,0.1,0.9} #1}}

\newcommand{\bracedtext}[1]{{\{\{\xspace#1\xspace\}\}}}

\newcommand{\beir}{\textsf{BEIR}\xspace}
\newcommand{\trecdl}{\textsf{TREC-DL}\xspace}
\newcommand{\mrtydi}{\textsf{Mr.TyDi}\xspace}

\newcommand{\ndcg}{\textsf{nDCG}\xspace}
\newcommand{\recall}{\textsf{Recall}\xspace}

\newcommand{\treccovid}{\textsf{TREC-COVID}\xspace}
\newcommand{\nfcorpus}{\textsf{NFCorpus}\xspace}
\newcommand{\signal}{\textsf{Signal-1M (RT)}\xspace}
\newcommand{\trecnews}{\textsf{TREC-NEWS}\xspace}
\newcommand{\robust}{\textsf{Robust04}\xspace}
\newcommand{\touche}{\textsf{Touch\'e-2020}\xspace}
\newcommand{\dbpedia}{\textsf{DBPedia}\xspace}
\newcommand{\scifact}{\textsf{SciFact}\xspace}

\newcommand{\gptturbo}{\textsf{GPT-3.5-Turbo}\xspace}
\newcommand{\gptfour}{\textsf{GPT-4}\xspace}

\newcommand{\bm}{\textsf{BM25}\xspace}
\newcommand{\bmce}{\textsf{BM25+CE}\xspace}
\newcommand{\monobert}{\textsf{monoBERT}\xspace}
\newcommand{\monot}{\textsf{monoT5}\xspace}
\newcommand{\rankgpt}{\textsf{RankGPT}\xspace}
\newcommand{\colbert}{\textsf{ColBERT}\xspace}
\newcommand{\doctquery}{\textsf{docT5query}\xspace}
\newcommand{\sparta}{\textsf{SPARTA}\xspace}
\newcommand{\dpr}{\textsf{DPR}\xspace}
\newcommand{\genq}{\textsf{GenQ}\xspace}
\newcommand{\pgator}{\textsf{Promptagator}\xspace}
\newcommand{\tart}{\textsf{TART-Rerank (FLAN-T5-XL)}\xspace}
\newcommand{\upr}{\textsf{UPR}\xspace}

\newcommand{\ance}{\textsf{ANCE}\xspace}
\newcommand{\tasb}{\textsf{TAS-B}\xspace}

\newcommand{\rrr}{\textsf{RRR}\xspace}

\newcommand{\rewriter}{\ensuremath{g}\xspace}
\newcommand{\retriever}{\ensuremath{f}\xspace}
\newcommand{\rw}{\rewriter}
\newcommand{\rt}{\retriever}
\newcommand{\reranker}{\ensuremath{h}\xspace}
\newcommand{\relm}{\ensuremath{\sigma}\xspace}
\newcommand{\lr}{\relm}
\newcommand{\gr}{\reranker}

\newcommand{\query}{\ensuremath{q}\xspace}
\newcommand{\numdocs}{\ensuremath{N}\xspace}
\newcommand{\numaug}{\ensuremath{N_{\text{aug}}}\xspace}
\newcommand{\maxrw}{\ensuremath{N_{\text{rw}}}\xspace}
\newcommand{\relt}{\ensuremath{\tau}\xspace}
\newcommand{\rwprompt}{\ensuremath{\pi}\xspace}
\newcommand{\docset}{\ensuremath{\cS}\xspace}
\newcommand{\retset}{\ensuremath{\cR}\xspace}
\newcommand{\filset}{\ensuremath{\cF}\xspace}

\newcommand{\doc}{\ensuremath{z}\xspace}

\newcommand{\qspace}{\ensuremath{\mathcal{Q}}\xspace}
\newcommand{\docspace}{\ensuremath{\mathcal{Z}}\xspace}

\newcommand{\winsize}{\ensuremath{w}\xspace}
\newcommand{\stepsize}{\ensuremath{s}\xspace}

\mdfdefinestyle{prompt}{
    outerlinewidth = 2,
    roundcorner = 1pt,
    leftmargin = 5,
    rightmargin = 5,
    backgroundcolor = gray!20,%
    outerlinecolor = blue!70!black,%
}

%% file: 1-intro.tex
\section{Introduction}\label{sec:intro}

We consider the information retrieval (IR) problem arising in search \citep{belkin03search, Ian08ir, dahiya21xml}, recommendations \citep{su09reco, covington16youtube, vemuri23xc}, and open-domain question-answering \citep{brill2002analysis, roberts-etal-2020-much, zhu21qasurvey}. Given an input query and a possibly large corpus of (text) documents, the goal is to retrieve relevant documents for the query. 
The retrieval problem is at least as old as Internet search~\citep{pinkerton1994finding,kobayashi2000information}. However, instead of the standard paradigm of training a new model for each retrieval task or domain, lately there has been a lot of attention on \textit{zero-shot} retrieval~\citep{beir,nguyen2640ms,mmarco}. In this setup, there is no access to training data from the target retrieval domain and the model is expected to generalize from its pre-trained data. 
Progress in zero-shot retrieval can be attributed 
to rich world knowledge implicit in the pre-trained model parameters of language models such as BERT~\citep{monobert, yang-etal-2019-end-end} 
and more recently generative models like GPT-3.5 \citep{chatgpt}. In particular, instruction-following abilities of large generative models have been shown to enable state-of-the-art accuracy on zero-shot benchmarks  such as \beir \citep{beir} and
\mrtydi \citep{mrtydi}.



Algorithms for retrieval typically contribute to one of the following stages in an end-to-end IR pipeline: 
\textbf{1) }\textit{rewrite}, augment the query with auxiliary information; \textbf{2)} \textit{retrieve}, fetch list of relevant documents;  and \textbf{3)} \textit{re-rank} the fetched list. There has been a steady line of research in the last few years that marry the generation capabilities of large language models with embedding-based retrieval models \citep{guu2020retrieval,lewis2020retrieval,singh2021end,izacard2022few,query2doc23,genfeedback23}, and replace one or more of these IR stages with learnable components. Two popular paradigms for information retrieval with language models are retrieval-augmented generation (RAG) and generation-augmented retrieval (GAR).\\
\textbf{1. RAG paradigm} \citep{chen2017reading,guu2020retrieval,lewis2020retrieval,singh2021end,izacard2022few} fetches (using a retrieval model) relevant documents from the corpus as context for the language model and then \textit{generates} an answer for the input query directly using the language model. \\
\textbf{2. GAR paradigm} \citep{query2doc23,genfeedback23} augments the input query using language models, and then uses a retrieval model to fetch the relevant documents from the corpus. 

A key challenge in these paradigms is obtaining a high-quality retrieval model for fetching documents during the first stage and a post-hoc re-ranking model to improve the precision of the final top-$k$ results. Dense retrieval techniques like \ance \citep{ance}
\citep{monobert} and \tasb \citep{tasb} suffer from poor precision, and fine-tuning the models is infeasible in the zero-shot setting. Recent results \citep{beir} show, somewhat surprisingly, that off-the-shelf sparse retrieval models like BM25 \citep{bm25} outperform dense counterparts when combined with generative language models.  More strikingly, for re-ranking, recent studies by \cite{sungpt23,qinllm23}, show promise for designing effective re-ranking strategies using LLMs like GPT-4 as a black-box. These studies, however, do not consider feedback between the three stages. For example, a good initial ordering of retrieved documents is crucial for the re-ranking to be effective and a good rewrite of the input query can improve the quality of retrieved documents.

\begin{figure*}
    \centering
    \includegraphics[width=0.99\textwidth]{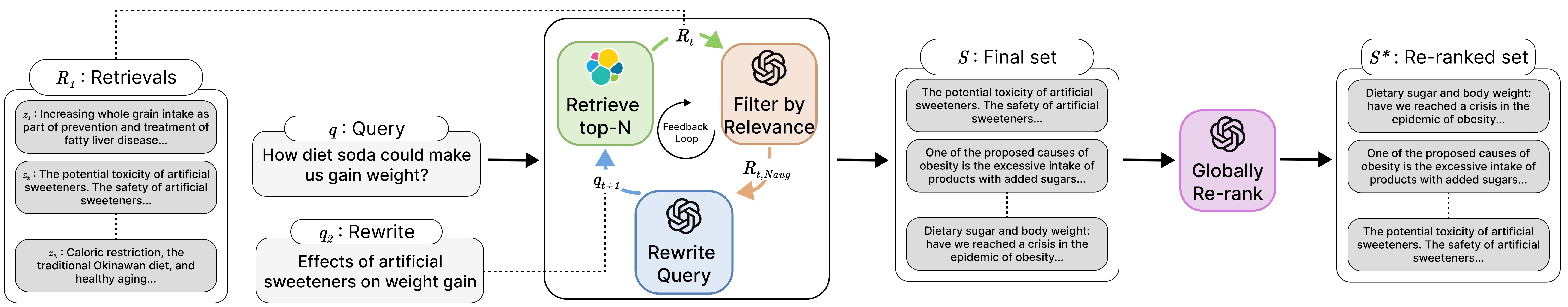}
    \caption{Proposed \rrr method for zero-shot Information Retrieval. We implement the rewrite, filtering, and re-rank stages (colored boxes) via a pre-trained LLMs (in our evaluations, we use \gptturbo and \gptfour models). For the retrieval stage, we use BM25. Details in Section \ref{sec:method}.}
    \label{fig:RRR}
\end{figure*}

In this work, we achieve the best of both the worlds, i.e., of GAR and RAG paradigms. We propose a novel GAR-meets-RAG formulation for zero-shot IR that incorporates a feedback loop of rewrite and retrieval stages. We design a simple and effective approach to IR, called \rrr (Rewrite-Retrieve-Rerank), that leverages pre-trained models to perform document retrieval, refinement, and query rewrite iteratively (Figure \ref{fig:RRR}). The key design principle is that the rewrite-retrieval stages improve the recall of the system and a final re-ranking stage improves the precision.  A key technical contribution in this work is a novel prompting strategy for the query rewrite stage which allows the rewriter to be aligned to the type of documents present in the unseen corpus.  

Our contributions are summarized below: \\
\textbf{(1)} We propose a novel GAR-meets-RAG recurrence formulation for the zero-shot IR problem, that uses a RAG model to produce query rewrite, which feeds into a GAR model for retrieval.  \\ 
\textbf{(2)} We design a simple, iterative algorithm for the proposed problem called \rrr that maximizes recall via rewrite-retrieve stages and precision via a final re-rank stage. \\
\textbf{(3)} We perform extensive evaluations and comparisons to SOTA techniques on two popular IR benchmarks. We establish new state-of-the-art \recall{}@100 and \ndcg{}@10 metrics on 6 out of 8 datasets in the BEIR benchmark, with up to 17\% relative gains over the previous best. 

%% file: 3-problem.tex
\section{Background and Notation}\label{sec:problem}
Given a query, our task is to retrieve \textit{relevant} documents from a large corpus, without using any training data specific to the domain.  For example, queries can be factual such as ``\textit{Can antioxidant-rich spices counteract the effects of a high-fat meal?}" from the \nfcorpus dataset or open-ended such as ``\textit{Should teachers be given tenure?}" from the \touche dataset. The ground-truth \textit{relevant} documents for these queries are online documents (e.g., scientific journal abstracts, tweets, articles) that contain information pertinent to the query\footnote{\url{https://pubmed.ncbi.nlm.nih.gov/21697300} is relevant for the \textit{antioxidant} query in the \beir benchmark.}. To fetch relevant documents for a query, a \textit{retrieval model} is trained that can optionally be improved using a \textit{relevance model} or \textit{generative model}, as described below.

\paragraph{Zero-shot IR Problem:} The input to the retrieval system are: (i) a query, denoted by $\query \in \qspace$, which is a sequence of tokens, and (ii) corpus of text documents $\docspace  \ni \doc$ that are indexed using standard techniques for retrieval \citep{bm25}; 
$|\docspace|$ can be in the order of millions. For evaluation, we have access to ground-truth relevance labels of the form $\langle \query, \docset^*=\{\doc^*_j, r^*_j\} \rangle$, where $|\docset^*|$ is typically very small (much smaller than $|\docspace|$). Here, $r^*_j > 0$ denotes an ordinal relevance score for the pair $q, \doc^*_j$.  However, in our zero-shot setup, the retrieval system does not have access to the relevance labels from the corpus at any point in time.  We seek a model that produces a ranked list of documents $\docset = \langle \doc_1, \doc_2, \dots, \doc_\numdocs \rangle$ for a given input query $\query$ such that the retrieval quality (measured as follows) is high. 

\paragraph{Metrics:} We focus on two metrics standard in information retrieval research \citep{beir,ance,sungpt23}: \\
\textbf{(1)} $\ndcg@k$ \citep{jarvelin2002cumulated} which is the standard metric of interest for ranking problems. For binary relevance feedback, $\ndcg@k$ is maximized for a query when (i) the relevant documents in $\docset$ are ranked above all the irrelevant documents in $\docset$, \textit{and} (ii) $|\docset \cap \docset^*|$ (prop. to Precision$@k$) is maximized.\\
\textbf{(2)} $\recall@k$ which measures the fraction of relevant documents retrieved for the query, i.e., $|\docset \cap \docset^*|/|\docset^*|$. 


\paragraph{Retrieval model:} A retrieval (or a \textit{f}etch) model $\rt:\qspace \to 2^\docspace$ maps a query to a (small) list of potentially relevant documents. Two popular retrieval models are (i) dense retrieval models that embed queries and documents in a vector space with dimensionality much smaller than the token vocabulary size. The embedding function is typically modeled via deep encoders and learned using relevance labels and a contrastive loss function \citep{dpr,ance}; (ii) sparse retrieval models, e.g., BM25  \citep{bm25}, on the other hand, use a simpler tf-idf \citep{salton1988term} scheme based on the relative frequency of words occurring in different documents. In either case, the retrieval model $\rt$ computes dot-product similarity scores between the query and the document embeddings and retrieves the top-$k$ most similar documents. 

\paragraph{Generative (language) model:} A language (or a \textit{g}enerative) model $\rw$ is a sequence-to-sequence (Transformer-based) model that produces output text (e.g., a query rewrite, a document, or an answer) conditioned on the input text (e.g., a query). In the RAG paradigm \citep{guu2020retrieval,yu2023generate}, $\rw$ takes a query and retrieved documents (using a retrieval model as above) as input and generates an answer as output. In the GAR paradigm \citep{nogueira2019document,query2doc23}, $\rw$ takes a query as input, and generates additional context such as query expansions, document expansions as long-form text, which is then used as input to the retrieval model. 

\paragraph{Relevance model:} A relevance model $\relm: \docspace \times \qspace \to \Real$ takes a query, document pair and computes a relevance score. In this work, we use ordinal relevance scores. Relevance models are employed in IR for multiple reasons, including (1) de-noising hard negatives for improved training of retrieval models \citep{qu2021rocketqa} as there can be spurious negatives (irrelevant documents) when sampled from a very large corpus, (2) as a filtering and/or re-ranking mechanism to improve retrieval performance \citep{nogueira2019multi,zhou2022towards}. In the supervised settings, relevance models are trained using bi-encoders or cross-encoders. Recent studies in various domains \citep{bai2022constitutional,kocmi2023large,zhuo2023large,liang2023holistic} show that large language models (as a black-box) can be used to design effective relevance models.

%% file: 4-method.tex
\section{Proposed Method: GAR meets RAG}
In this section, we first introduce a novel formulation for the IR problem, and then present our iterative algorithm for the zero-shot setting that leverages a pipeline of pre-trained language models.
\subsection{Formulation}
\label{sec:formulation}
We formulate the problem of retrieving top-$\numdocs$ relevant documents for an input query $\query$ as a composition of GAR (i.e., first generate, then retrieve) and RAG (first retrieve, then generate) models. For clarity, we suppress  additional arguments that generative and retrieval models typically need (the number of documents to retrieve, prompt construction, etc.) here, and give details in the next subsection.

Consider the GAR paradigm \citep{nogueira2019document,query2doc23,genfeedback23} first.
Recall that, here, one uses a language model $\rw$ to generate augmented query context given $\query$, which then becomes the input for the retrieval model $\retriever$ along with, optionally, the original query. We can write the GAR paradigm as the composition $\docset = \retriever\big(\query; \rw(\query)\big)$. The success of GAR paradigm hinges on (i) having a high-quality retrieval model $\retriever$, and (ii) the quality of the additional context produced by the model $\rw$.  


Now, consider the RAG paradigm. Recall that, here, one uses the retrieval model $\retriever$ first to fetch potentially relevant documents $\docset$; which then becomes the input to the language model $\rw$ to produce answer for the query directly. In other words, we can write the RAG paradigm as the composition $\tilde{\doc} = \rw\big(\query; \retriever(\query)\big)$; where tilde on $\doc$ denotes that it is a generated answer and not a ground-truth document.

We make two simple but key observations: (1) we can use the RAG model as a way to provide the context that is crucial for the GAR model; (2) similarly, we can use the GAR model itself as the retriever model for the RAG model. This motivates us to formulate the retrieval problem as the following GAR-meets-RAG recurrence:
\begin{align}
\query_1 = \query&, \ \ \docset_0 = \{\}, \nonumber \\
\docset_t &= \docset_{t-1} \oplus \retriever(\query_t), \\
\query_{t+1} := \tilde{\doc}_{t+1} &= \rw\big(\{\query_{t-i}; \retriever(\query_{t-i})\}_{i=0}^{t-1}\big),
\end{align}

where $\oplus$ denotes a suitable list merge operator (to be defined shortly) that ensures that the output documents $\docset_\infty$ are sorted (we discuss the termination criteria in Section \ref{sec:method}).
Recall (from Section \ref{sec:problem}) that $\retriever$ returns a list of documents sorted by the retrieval scores (e.g., dot-product similarity between query and document embeddings). 

A few remarks are in order: (a) we use the RAG model, i.e., (2) above, to generate a \textit{query rewrite} or \textit{a reformulation} $\query_{t+1} \in \qspace$ as the output $\tilde{\doc}_{t+1}$. 
Using query rewrites for search and information retrieval is a popular technique \citep{abdul2004umass,query2doc23,generativeprf23}. The novelty in this formulation is that we are employing a RAG model to produce the rewrite; (b) we use the query rewrite as the input for the retrieval model in the spirit of GAR paradigm in (1); (c) we can adapt and improve the generation quality and the retrieval results via the feedback loop implicit in the recurrence. 

\textbf{Challenges:} The recurrence formulation as stated presents multiple design challenges.

\textbf{Firstly}, the right way to define $\oplus$ for merging lists in (1) at each iteration is unclear. An immediate idea is to use similarity scores of \retriever itself to rank and merge the lists. But, this is problematic as the scores are not calibrated across queries, i.e., the scores of $(\query,\cdot)$ and $(\query^\prime,\cdot)$, for queries $\query \neq \query^\prime$, are not comparable. 
Calibrating the scores needs access to validation data from the target domain, which we lack in the zero-shot setting. 

\textbf{Secondly}, since we want the retrieval system to output top-$\numdocs$ relevant documents to the query, we have the constraint $|\docset_\infty| = \numdocs$. So, we need to appropriately control the size of $|\docset_t|$ in (1), while also trying to maximize the performance metrics of interest (stated in Section \ref{sec:problem}). 

\textbf{Finally}, a poor query rewrite in one iteration could corrupt the subsequent retrievals, which could in turn derail the subsequent query rewrite, and so on. 
Thus, we need to ensure the formulation is less brittle.

We solve the above challenges via introducing a relevance model $\relm$ (discussed in Section \ref{sec:problem}). \\ 
\textbf{(i)} We define $\oplus$ as the merge operation using the scores of the documents for the \textit{original input query} $\sigma (q, \cdot)$, and let $\Pi_{\sigma(q, \cdot)}$ denote the ranking induced by the corresponding scores (we empirically validate this design choice in Section \ref{sec:results}). \\
\textbf{(ii)}  We also use $\sigma (q, \cdot)$ to filter poor quality retrievals in recurrence (1). Together with \textbf{(i)}, it helps ensure that the intermediate retrievals $\docset_t$ are also highly relevant to the original query which in turn helps maximize the recall metric, subject to $|\docset_\infty| = N$ constraint. \\
\textbf{(iii)} 
To ensure subsequent rewrites do not deviate too much from the original intent of the query $\query$, and to make the formulation less brittle
, we include only a top few (a configurable number) highly relevant retrievals of $\retriever$, denoted by $\retriever_\text{trim}$ in (2). 

The modified recurrence is given by:
\begin{align}
\label{eqn:formulation}
\query_1 = \query&, \ \ \docset_0 = \{\}, \nonumber \\
\docset_t &= \docset_{t-1} \oplus \Pi_{\sigma(q, \cdot)}\big(\retriever(\query_t)\big), \\
\query_{t+1} := \tilde{\doc}_{t+1} &= \rw\big(\big\{\query_{t-i};  \retriever_\text{trim}(\query_{t-i})\big\}_{i=0}^{t-1}\big).\nonumber 
\end{align}

Next, we give a formal algorithm, discuss details of design choices and implementation.

\subsection{Algorithm: \rrr}
\label{sec:method}
\input{4.0-algorithm}
We give the procedure for implementing the recurrences \eqref{eqn:formulation} in Algorithm \ref{alg:rrr} titled \rrr. Besides a query rewriter model \rewriter, a retrieval model \retriever, a relevance model $\lr$, that are needed in the recurrences, the algorithm also uses a re-ranker model $\gr$ at the end (discussed shortly). In our evaluations, we use black-box LLMs for \rewriter, $\lr$, and $\gr$ (all with zero-shot prompts); and the standard sparse retrieval model BM25 for \retriever. Prompt templates are provided in Appendix \ref{app:prompts}.

The algorithm takes as input (i) total number of output documents \numdocs, (ii) maximum number of query rewrites 
(iii) maximum number of retrievals \numaug to augment the prompt of rewriter \rewriter with (i.e., to implement $\retriever_\text{trim}$ in (3)), and (iv) a threshold \relt for the scores $\relm(\query, \cdot)$, to remove spurious documents (false positives) at each iteration.

The algorithm starts with an initial prompt $\rwprompt_0$ for the rewriter LLM \rewriter (given in the Appendix) and empty set \docset of retrieved documents. At iteration $t$:

\textbf{Stage 1. Retrieve and filter:} First, we retrieve a list of $\numdocs$ documents $\retset_t$ for the query $\query_t$ using the retrieval model \retriever. Consider the input query 
``\textit{How diet soda could make us gain weight?}"
shown in Figure \ref{fig:RRR} at the first iteration: From $\retset_1$ shown in Figure \ref{fig:RRR}, we observe that while the top document is likely relevant, there are some spurious candidates such as ``\textit{Increasing whole grain intake as part of prevention and treatment of fatty liver disease...}" and ``\textit{Caloric restriction, the traditional Okinawan diet, and healthy aging...}". 
To prune such false positives, we obtain relevance scores $\relm(\query,\doc)$ for each retrieved document $z \in \retset_t$ using the relevance model $\relm$ (the prompt template for the $\relm$ LLM is provided in Appendix \ref{app:prompts}); the higher the score, the more relevant the document is to the original query $\query$. We add the documents in $\retset_t$ that clear the relevance threshold \relt, denoted by $\filset_t$, to the output set $\docset$.  



\textbf{Stage 2. Rewrite:} We take the top-$\numaug$ documents from $\retset_{t}$, denoted by $\retset_{t,\numaug}$. This list, together with the (rewritten) query $\query_t$, is appended to the prompt of the rewrite model \rewriter (see Appendix for the \Call{Append}{} implementation) to derive $\rwprompt_{t}$ and generate a new query rewrite $\query_{t+1}$. This trimmed set $\retset_{t,\numaug}$ also helps work with limited prompt sizes of LLMs (e.g., GPT-3.5-turbo has a limit of 4097 tokens) when the documents are large. In Figure \ref{fig:RRR}, given query $\query_1$ and $\retset_{1, \numaug}$, the rewriter \rewriter generates a plausible rewrite $\query_2$ ``\textit{Effects of artificial sweeteners on weight gain}". 

We iterate the two stages until either $|\docset|$ exceeds $\numdocs$ or the number of rewrites exceeds \maxrw. We show empirically that the resulting $\docset$ achieves high recall.  

\textbf{Stage 3. Re-rank:} In the final stage, we first rank the documents $\docset$ by their relevance scores w.r.t. input query $\query$, $\relm(\query, \doc), \doc \in \docset$. This becomes the initial ordering for the LLM-based re-ranker model \gr, which we implement using the technique of \cite{sungpt23}. The algorithm outputs the re-ranked list $\docset^* = \gr(\docset)$ for the input query $\query$.

%% file: 4.0-algorithm.tex
\begin{algorithm*}[h]

\small
\caption{\rrr: Rewrite, Retrieve, Re-rank}\label{alg:rrr}
\begin{algorithmic}[1]

\State \textbf{Input:} query \query, corpus \docspace, rewriter \rewriter, retriever \retriever, relevance model \relm, relevance threshold \relt, re-ranker \reranker, \#docs to retrieve \numdocs, \#retrievals to augment in the rewriter prompt \numaug, max \#rewrites \maxrw
\State \textbf{Initialize}: $\query_1 \leftarrow \query$, output document set $\docset \leftarrow \{\}$, rewrite prompt $\rwprompt_0(q)$

\For{$t \gets 1, \dots, \maxrw$}

\circled{1} \stc{\textbf{R}etrieve and filter} 
    
    \State Retrieve \numdocs documents from \docspace, $\retset_t \leftarrow \retriever(\query_t)$, for query $\query_t$ using the retrieval model \retriever
    
    \State Obtain relevance scores $\relm(z,\query), z \in \retset_t$ from the relevance model \relm
    
    \State Get filtered document set $\filset_t \leftarrow \lbrace \doc \in \retset_t\ |\ \relm(\query, \doc) > \relt \rbrace$
    
    \State Add to \docset, i.e., $\docset \leftarrow \docset \cup \filset_t$ 

    \If{$|\docset| \geq \numdocs$}
        \State \textbf{break}
    \EndIf

\circled{2} \stc{\textbf{R}ewrite} 
    \State Take top \numaug documents $\retset_{t,\numaug}$ from $\retset_t$ (using retriever scores in Step 4)
    \State Add $\query_t$ and $\retset_{t,\numaug}$ to the prompt, i.e., $\rwprompt_{t} \leftarrow \Call{Append}{\rwprompt_{t-1}, \query_t, \retset_{t,\numaug}}$
    
    \State Generate new rewrite $\query_{t+1} = \rewriter(\query_t;\rwprompt_{t})$

\EndFor

\State Order documents by relevance scores, i.e., $\docset \leftarrow \Pi_{\sigma(\query, \cdot)}\big(\docset)$

\hspace{-1cm}\circled{$3$} \stc{\textbf{R}e-rank using LLM-based \reranker} 
\State \textbf{return} $\cS^* = \reranker(\docset)$ 

\end{algorithmic}
\end{algorithm*}

%% file: 5.5-table_beir.tex
\begin{table*}[!ht]
\caption{Retrieval performance (\ndcg{}@10) on \beir{} datasets. Dataset-wise best score is marked in \textbf{bold} and the second best is \underline{underlined}.}
\label{tab:beirndcg}
\resizebox{1\textwidth}{!}{%
\centering
\begin{tabular}{c c c c c c c c c c}
    \toprule
    Method & \treccovid & \nfcorpus & \signal & \trecnews & \robust & \touche & \dbpedia & \scifact \\ 
    \toprule
    \bm & 59.5 & 30.8 & \underline{33.0} & 39.5 & 40.7 & \textbf{44.2} & 31.8 & 67.9 \\ 
    \dpr & 33.2 & 18.9 & 15.5 & 16.1 & 25.2 & 13.1 & 26.3 & 31.8 \\ 
    \ance & 65.4 & 23.7 & 24.9 & 38.2 & 39.2 & 24.0 & 28.1 & 50.7 \\ 
    \tasb & 48.1 & 31.9 & 28.9 & 37.7 & 42.7 & 16.2 & 38.4 & 64.3 \\ 
    \monot (3B) & 80.7 & \underline{39.0} & 32.6 & 48.5 & 56.7 & 32.4 & 44.4 & \underline{76.6} \\ 
    \pgator++(few-shot) & 76.2 & 37.0 & - & - & - & 38.1 & 43.4 & 73.1 \\ 
    \rankgpt & \underline{85.5} & 38.5 & \textbf{34.4} & \underline{52.9} & \underline{57.6} & \underline{38.6} & \underline{47.1} & 75.0 \\ 
    \rrr (this work) & \textbf{86.4} & \textbf{39.9} & 29.8 & \textbf{53.6} & \textbf{67.4} & 29.8 & \textbf{51.0} & \textbf{77.2} \\ 
    \toprule
\end{tabular}
}
\end{table*}

\begin{table*}[!ht]
\caption{Retrieval performance (\recall{}@100) on \beir{} datasets. For \treccovid, capped \recall{}@100 is used. Dataset-wise best score is marked in \textbf{bold} and the second best is \underline{underlined}.}
\label{tab:beirrecall}
\resizebox{1\textwidth}{!}{%
\centering
\begin{tabular}{c c c c c c c c c c}
    \toprule
    Method & \treccovid & \nfcorpus & \signal & \trecnews & \robust & \touche & \dbpedia & \scifact \\ 
    \toprule
    \bm  &  \underline{49.8}  & 24.6 &  \textbf{37.0}  &  \underline{44.7}  &  \underline{37.5}  &  \textbf{58.2}  & 46.8 &  \underline{92.5} \\ 
    \dpr  & 21.2 & 20.8 & 16.2 & 21.5 & 21.1 & 30.1 & 34.9 & 72.7 \\ 
    \ance  & 45.7 & 23.2 & 23.9 & 39.8 & 27.4 & 45.8 & 31.9 & 81.6 \\ 
    \tasb  & 38.7 &  \underline{28.0}  & 30.4 & 41.8 & 33.1 & 43.1 &  \underline{49.9}  & 89.1 \\ 
    \monot (3B)  &  \underline{49.8}  & 24.6 &  \textbf{37.0}  &  \underline{44.7}  &  \underline{37.5}  &  \textbf{58.2}  & 46.8 &  \underline{92.5} \\ 
    \rankgpt  &  \underline{49.8}  & 24.6 &  \textbf{37.0}  &  \underline{44.7}  &  \underline{37.5}  &  \textbf{58.2}  & 46.8 &  \underline{92.5} \\ 
    \rrr (this work)  &  \textbf{54.8}  &  \textbf{32.4}  &  \underline{32.4}  &  \textbf{51.6}  &  \textbf{45.4}   &  \underline{52.2}  &  \textbf{55.0}  &  \textbf{94.3} \\ 


    \toprule
\end{tabular}
}
\end{table*}

%% file: 5.6-table_trecdl.tex
\begin{table}
\caption{Performance (\ndcg{}$@k$) on the \trecdl{}20 dataset. Best score is marked in \textbf{bold} and the second best is \underline{underlined}.}
\label{tab:trecdl}
\centering
\begin{tabular}{c | c c c}
    \toprule
    Method & \ndcg@1 & \ndcg@5 & \ndcg@10 \\
    \toprule
    \bm  & 57.7 & 50.7 & 48.0 \\
    \monobert (340M)  & 78.7 & 70.7 & 67.3 \\
    \monot (220M)  & 77.5 & 69.4 & 67.0 \\
    \monot (3B)  &  \textbf{80.3}  & 72.3 & 68.9 \\
    \upr  & 63.2 & 59.4 & 56.0 \\
    \rankgpt  & 78.4 &  \underline{74.1}  &  \underline{70.6} \\
    \rrr (this work)  &  \underline{79.6}  &  \textbf{76.0}  &  \textbf{72.3} \\
    \toprule
\end{tabular}
\end{table}

    

%% file: 5-evalsetup.tex
\section{Evaluation Setup}
\label{sec:evalsetup}
In this section, we give details of datasets, baselines, metrics, and implemention, before presenting evaluation results in the subsequent section.

\paragraph{(A) Datasets and metrics:}
We present quantitative evaluations of \rrr on two standard IR benchmarks: 
\beir(\cite{beir}) and \trecdl(\cite{trecdl19, trecdl20}). Due to resource constraints (LLM calls), we perform evaluations on subsets of these benchmarks as listed below (each input query costs approximately 1USD to complete; cost breakdown in Appendix \ref{app:cost}).

\textbf{\beir} is a benchmark for comprehensive zero-shot evaluation of models across a diverse spectrum of IR tasks. We take 8 datasets with relatively small test sets (listed in Table \ref{tab:beirndcg}) 
out the 18 total available. 

\textbf{\trecdl} is a dedicated deep-learning track within the Text Retrieval Conference (TREC) encompassing document retrieval and passage retrieval tasks. We report results on the passage retrieval dataset \trecdl{}20, with 54 queries and 8.8M documents. For hyperparameter tuning, we use the \trecdl{}19 dataset with 43 queries; see \textbf{(D)} below.

We report \ndcg{}@10 and \recall{}@100 metrics.

\paragraph{(B) Compared techniques:} We compare with several zero-shot IR methods: dense passage retrieval \dpr \citep{dpr}, \ance \citep{ance}, \tasb \citep{tasb}, sparse retrieval \bm \citep{bm25}), document re-ranking \monot \citep{monot}, \rankgpt \citep{sungpt23}, and data generation/augmentation \pgator \citep{pgator22}. For all the baselines, we quote results from \cite{beir} and \cite{sungpt23} computed in identical setting.

\paragraph{(C) Implementation details:}
We run all our experiments on a single machine with an A-100 GPU, 24 CPU cores and 220 GB of memory. We use \gptturbo (\cite{chatgpt}), with a token limit of 4097, and \gptfour(\cite{gpt4}), with a token limit of 8192.

The configurations for different components of the \rrr algorithm are listed in Table \ref{tab:model_configurations}. For the retriever \retriever, we use the \bm model from the \texttt{Pyserini} \citep{pyserini} package. The re-ranker \reranker follows the two-step process described in the \rankgpt paper \citep{sungpt23}: first, re-rank \numdocs documents in $\docset$ (\textit{after} Step 15 of Algorithm \ref{alg:rrr}) using \gptturbo; then, further re-rank the top 30 documents using \gptfour. 

\begin{table}
\caption{\rrr configuration ({\footnotesize Temperature=0 for LLMs}).}
\label{tab:model_configurations}
\centering
\begin{tabular}{c c c}
    \toprule
    Component & Model & Details \\
    \toprule
    Rewriter \rewriter & \gptfour& token limit 20 \\
    Retriever \retriever & \bm &  \cite{pyserini} \\
    Relevance \relm & \gptturbo & score $\in [1,2,\dots,5]$ \\
    Re-ranker \reranker & \gptturbo + \gptfour & \cite{sungpt23} \\
    \toprule
\end{tabular}
\end{table}

\paragraph{(D) Hyperparameter selection:}
We perform hyperparameter tuning on the \trecdl{}19 dataset. First, we tune the number of rewrites \maxrw and the relevance threshold \relt by optimizing \recall{}@100. Secondly, to optimize \ndcg{}@10, we tune the sliding window constants \winsize{} (window size) and \stepsize{} (step size) in the re-ranker \reranker. The best values for these parameters are \maxrw = 5, \relt = 1 (i.e., prune the least scoring ones), \winsize = 10 and \stepsize = 5. See Appendix \ref{app:hparam_tuning} for details.





%% file: 5-results.tex
\vspace{-0.2cm}
\section{Results}
\label{sec:results}
\vspace{-0.4cm}

We focus on the following questions in our evaluations:\\
\textbf{(1) End-to-end performance:} How effective is \rrr{} compared to baseline and SOTA zero-shot IR techniques on the benchmarks introduced in Section \ref{sec:evalsetup}?\\
\textbf{(2) Effectiveness of feedback loop:} How does \rrr{} perform with more rewrites?\\
\textbf{(3) Ablative study:} How do the design choices in the proposed method impact performance?

\subsection{End-to-end performance}
\vspace{-0.1cm}
We show the performance metrics for the compared methods on the \beir datasets in Tables \ref{tab:beirndcg} and \ref{tab:beirrecall}. Our method outperforms all the baselines and SOTA zero-shot IR techniques on 6 out of 8 datasets, on both \ndcg{}@10 and \recall{}@100 metrics, achieving up to 17\% relative gain in the nDCG metric over state-of-the-art RankGPT. On the two datasets where \rrr{} is not performing well, the baseline \bm{} retrieval model outperforms nearly all of the sophisticated techniques. On the \trecdl{}20 dataset, shown in Table \ref{tab:trecdl}, we outperform all the methods on \ndcg{}@10 and \ndcg{}@5 metrics, while being competitive on \ndcg{}@1.

These results demonstrate the effectiveness of our method in the zero-shot setting; \rrr outperforming methods that also utilize powerful LLMs for IR is particularly significant. For instance, \pgator utilizes LLMs to generate queries from a corpus, and subsequently trains a dense retriever. Similarly, \monot and \rankgpt utilize LLMs (like \gptfour and \gptturbo) for re-ranking results.

\subsection{Effectiveness of feedback loop}

We wish to evaluate whether our method benefits from (a) retriever feedback (in recurrence Eqn. (3)), and (b) multiple feedback cycles. For this, we conduct experiments on \trecdl{}20, \trecdl{}19, and \treccovid. 

To address (a), we re-run experiments while omitting the addition of retrieved results to the re-writer prompt $\rwprompt_t$ (in Step 12 of Algorithm \ref{alg:rrr}). This effectively cuts off feedback from the retriever \retriever to the re-writer \rewriter. The results presented in Table \ref{tab:feedback_effect} illustrate that feedback is pivotal in order to improve recall, significantly so in \trecdl{19} and \trecdl{20} datasets. 

For (b), we vary the maximum number of rewrites \maxrw = \{1,3,5\} allowed for the rewriter \rewriter. The results in Table \ref{tab:maxrw_v_recall} show a discernible trend: \recall{}@100 improves as the number of rewrites increases. 

\subsection{Ablative study}


Here, we study the effects of design choices in \rrr (discussed in Sections \ref{sec:formulation} and \ref{sec:method}). 

\textbf{(i)} \textit{Is the final re-ranker \emph{\reranker} necessary}? Can we instead only rely on the relevance scores obtained using $\relm$ to re-rank \docset? Our results in Table \ref{tab:rerank_effect} show that the final re-ranker \reranker is critical for improving \ndcg{}@10. We hypothesize that this is because the discretized scores given by $\relm$ are insufficient for fine-grained ranking.

\textbf{(ii)} \textit{Should the relevance of a document be assessed in relation to the original query $\query_1$ (as in Algorithm \ref{alg:rrr}) or the rewrite $\query_{t}$ from which the document was retrieved}? Our results, as presented in Table \ref{tab:rel_query_rewrite}, indicate that assessing relevance should be based on the original query instead of the rewrite.

\textbf{(iii)} \textit{What is better feedback for the re-writer \rewriter (in Step 11 of Algorithm \ref{alg:rrr})}? We compare the use of top results based on retriever \retriever (\bm) scores with those  based on relevance (\relm) scores. Our findings in Table \ref{tab:feedback_source} suggest a slight advantage in using relevance scores to generate rewrites. However, because we use \trecdl{}19 for all design choices and hyperparameter tuning, we choose retriever-based feedback for our experiments.





%% file: 5-ablation.tex
\begin{table}
\caption{Effect of feedback from \retriever to \rewriter in \rrr. Scores reported are \recall{}@100.}
\label{tab:feedback_effect}
\centering
\footnotesize
\begin{tabular}{c | c | c}
    \toprule
    Dataset & With feedback & Without feedback \\
    \toprule
    \trecdl{}19 & \textbf{54.0} & 51.1\\
    \trecdl{}20 & \textbf{58.6} & 56.8 \\
    \treccovid & 13.0 & \textbf{13.2} \\
    \toprule
\end{tabular}
\end{table}

\begin{table}
\caption{\recall{}@100 on \trecdl{}20, \trecdl{}19 and \treccovid with varying number of rewrites \maxrw.}
\label{tab:maxrw_v_recall}
\centering
\footnotesize
\begin{tabular}{c | c c c}
    \toprule
    Dataset & \multicolumn{3}{c}{\maxrw} \\
    \cline{2-4}
    & 1 & 3 & 5 \\
    \toprule
    \trecdl19 & 49.7 & 52.6 & \textbf{54.0} \\
    \trecdl20 & 55.1 & 57.9 & \textbf{58.6} \\
    \treccovid & 12.5 & \textbf{13.0} & \textbf{13.0} \\
    \toprule
\end{tabular}
\end{table}

\begin{table}[t]
\caption{Effect of re-ranking (in Step 6 of Algorithm \ref{alg:rrr}). Scores reported are \ndcg{}@10.}
\label{tab:rerank_effect}
\centering
\footnotesize
\begin{tabular}{c | c | c}
    \toprule
    Dataset & With reranking & Without reranking \\
    \toprule
    \trecdl{}19 & \textbf{73.9} & 51.3\\
    \trecdl{}20 & \textbf{72.3} & 42.0 \\
    \treccovid & \textbf{86.4} & 66.8 \\
    \toprule
\end{tabular}
\end{table}

\begin{table}[t]
\caption{Relevance to original query $\query$ (as in Algorithm \ref{alg:rrr}) or rewrite $\query_{t}$? Scores reported are \recall{}@100.}
\label{tab:rel_query_rewrite}
\centering
\footnotesize
\begin{tabular}{c | c | c}
    \toprule
    Dataset & Relevance to $\query_1 = \query$ & Relevance to $\query_{t}$ \\
    \toprule
    \trecdl{}19 & \textbf{54.0} & 53.6 \\
    \trecdl{}20 & \textbf{58.6} & 58.0 \\
    \treccovid & \textbf{13.0} & 12.8 \\
    \toprule
\end{tabular}
\end{table}

\begin{table}[t]
\caption{Feedback from retriever (\bm) or relevance (\relm) model?  Scores reported are \recall{}@100.}
\label{tab:feedback_source}
\centering
\footnotesize
\begin{tabular}{c | c | c}
    \toprule
    Dataset & Feedback using \bm & Feedback using \relm  \\
    \toprule
    \trecdl{}19 & \textbf{54.0} & 53.1 \\
    \trecdl{}20 & 58.6 & \textbf{59.4 }\\
    \treccovid & 13.0 & \textbf{13.1} \\
    \toprule
\end{tabular}
\end{table}

%% file: 2-related.tex
\section{Related Work}\label{sec:related}
Language models have found increasing applications in Information Retrieval (IR) over the past few years. In this section, we highlight recent advances that leverage the in-depth world knowledge of LLMs (implicit in their pre-trained parameters) with retrieval components (both non-parametric and parametric) that have access to external memory, at different stages of IR.

\subsection{Retrieval Augmented Generation (RAG)}
RAG paradigm has been applied extensively in open-domain Question-Answering. \cite{brill2002analysis} use an N-gram tiling technique as \rewriter; \cite{chen2017reading} use an answer span prediction model as \rewriter, while the retrieval components in both these cases are black-box. On the other hand, recent systems employ learnable retrieval components: \cite{lee2019latent} use BERT-style encoders for both \retriever and \rewriter and learn them jointly; \cite{guu2020retrieval,lewis2020retrieval,singh2021end,izacard2022few} extend it to jointly pre-training or fine-tuning LMs as \rewriter and Transformer-based encoders as \retriever. Most recently, \cite{yu2023generate} eschew the retriever component altogether, and propose a \textit{generate-then-read} paradigm to first synthesize context (i.e., relevant documents) instead of retrieving from a corpus, and then generate the final answer based on the input query and the context. RAG has also been applied in code generation, e.g., \cite{parvez2021retrieval}. 

In contrast to most of the aforementioned approaches, we work at the intersection of the (a) zero-shot setting with no access to training data from the test domain; and (b) top-$k$ retrieval setting where the performance is measured via ranking metrics, instead of exact-match or Rouge-L based metrics typically used with \textit{generated} answers in the RAG paradigm. 

\subsection{Generation Augmented Retrieval (GAR)}
Query expansion techniques generate additional context or pseudo-documents using $\rw$ \citep{query2doc23, jagerman23, shen23, feng23knowledge}. Query rewriting or reformulation techniques, that we incorporate in \rrr, attempt to generate alternate versions of query $\query$ using different prompting strategies for $\rw$ \citep{hyde22,conversationsearch23}. \cite{nogueira2019document} expand documents instead with queries relevant to $\query$. Generative-relevance feedback techniques \citep{genfeedback23, generativeprf23, grm23} use $\rw$ to generate long-form text as first-pass retrieved documents $\doc$, to then seed the second-pass retrieval with $\query$ and $\doc$. Related line of work is learning improved dense representations for queries using pseudo-relevance feedback \citep{yu2021improving,wang2021pseudo}.

Unlike aforementioned approaches, \rrr iteratively improves the rewriting and the retrieving stages.  

\subsection{Language Models for Re-ranking}
A recent line of work leverages pre-trained LMs to re-rank the outputs of baseline retrievers using (a) novel prompting strategies that elicit pairwise preferences \citep{qinllm23}, or (b) sliding-window techniques \citep{sungpt23} that re-rank only a small window of retrieved documents at a time, bubbling up the entire retrieved set gradually, or (c) the likelihood of reconstructing the query conditioned on the retrieved documents \citep{upr}. However, these approaches rely on a good initial ordering of the retrieved documents to be successful. Our empirical evaluations show that our algorithm ensures recall as well as a good initial ordering of documents. 

\subsection{Zero-Shot IR}
Improving retrieval performance in the zero-shot setting is an active area of research \citep{quill22,dragon23}. \cite{inparsv1,inparsv2,pgator22} use LLMs with a few-shot prompt to synthesize relevant queries or documents, and use the resulting synthentic paired data to train task-specific retrievers. \cite{sachan23art} first retrieve an initial set of documents $\{\doc\}$ given a query $\query$, induce soft-labels on $\langle \query, \{z\} \rangle$ via trying to reconstruct $\query$ given $\{z\}$, and subsequently improve the retriever using the soft-labeled data.

%% file: 6-conclusion.tex
\section{Concluding Remarks}\label{sec:conclusion}
We propose a novel zero-shot recurrence formulation for IR that iteratively improves RAG and GAR models, the two increasingly-adopted paradigms for IR with language models. We devise a three-stage IR pipeline that improves recall in the first two (rewrite-retrieve) stages and precision in the final (re-ranking) stage. Our method \rrr{} achieves state-of-the-art retrieval performance on several IR datasets. 
\paragraph{Limitations and Future Work.} Our evaluations do not control for the model sizes of the compared methods: some baselines use simple models, while some, including ours, use powerful LLMs. The trade-off between the gains in retrieval performance and the inference cost depends on the application scenario. However, the formulation developed in Section \ref{sec:formulation} is fairly general, and we believe we can learn much smaller models via distillation techniques for both relevance and rewrite stages. An open question is designing compact yet effective models for re-ranking in the zero-shot setting, that improve over and beyond what retrieval/relevance models can achieve.

%% file: 7-appendix.tex
\setcounter{table}{0}
\renewcommand{\thetable}{A\arabic{table}}

\section*{Appendix}
\section{Prompts}\label{app:prompts}

\subsection{Rewrite Prompt}

\begin{mdframed}[style=prompt]
\textbf{system}:\newline
You are an AI assistant that helps people find information.

\textbf{user}:\newline
I am using a search engine to find relevant documents related to the given TOPIC. The search engine doesn't work very well.\newline
I will give you the top search results for various QUERIES that I tried.\newline
You should suggest me other topics that I should search in order to find more interesting documents relevant to the TOPIC.

Since the search engine mostly does lexical matching, it could be weak in retrieving documents containing some words. Use those words to improve the overall search quality. 

Also, use your own knowledge and understanding of the TOPIC to generate rewrites related to topics which might not be present in the retrieved documents.

Enclose the answer in $<$Rewrite$>$$<$/Rewrite$>$. Do not give any explanation.

TOPIC: \bracedtext{query \query}

QUERY \#1: \bracedtext{rewrite $\query_1$}

TOP RESULTS: 

1. \bracedtext{Filtered Document 1 $\filset_{1, 1}$}

2. \bracedtext{Filtered Document 2 $\filset_{1, 2}$}

3. \bracedtext{Filtered Document 1 $\filset_{1, 3}$}

QUERY \#2: \bracedtext{rewrite $\query_2$}

TOP RESULTS: 

1. \bracedtext{Filtered Document 1 $\filset_{2, 1}$}

2. \bracedtext{Filtered Document 1 $\filset_{2, 2}$}

3. \bracedtext{Filtered Document 1 $\filset_{2, 3}$}
\end{mdframed}

\subsection{Relevance Prompt}
\begin{mdframed}[style=prompt]
\textbf{system}:\newline
You are an AI assistant that helps people find information.

\textbf{user}:\newline
Given a QUERY and a DOCUMENT, score the DOCUMENT on a scale of 1(least relevant to QUERY) to 5(most relevant to QUERY).\newline
Enclose the answer in $<$Score$>$$<$/Score$>$. For instance if you think the score should be 4, then answer
$<$Score$>$4$<$/Score$>$. Do not give any explanation.

QUERY: \bracedtext{rewrite $\query_i$}

DOCUMENT: \bracedtext{Retrieved document $\retset_j$}
\end{mdframed}

\clearpage

\subsection{Re-ranking Prompt}
The re-ranking prompt is the same as in RankGPT (\cite{sungpt23}).
\begin{mdframed}[style=prompt]
\textbf{system}:\newline
You are RankGPT, an intelligent assistant that can rank passages based on their relevancy to the query.

\textbf{user}:\newline
I will provide you with \bracedtext{window size} passages, each indicated by number identifier []. 
Rank the passages based on their relevance to query: \bracedtext{\query}

\textbf{assistant}:\newline
Okay, please provide the passages.

\textbf{user}:\newline
[1] \bracedtext{Document 1 $\docset^\prime_1$}

\textbf{assistant}:\newline
Received passage [1]

\textbf{user}:\newline
[2] \bracedtext{Document 2 $\docset^\prime_2$}

\textbf{assistant}:\newline
Received passage [2]

(more passages...)

\textbf{user}:\newline
Rank the \bracedtext{window size} passages above based on their relevance to the search query. The passages should be listed in descending order using identifiers. The most relevant passages should be listed first. The output format should be [] $>$ [], e.g., [1] $>$ [2]. Only response the ranking results, do not say any word or explain.
\end{mdframed}

\section{Cost Breakdown}\label{app:cost}
In our experiments, the rewriter uses \gptfour, the relevance model \relm uses \gptturbo, and the re-ranker \reranker uses both \gptturbo and \gptfour.

Following is a breakdown of the number of queries for each stage of our \rrr algorithm:

\textbf{Retrieve and Filter:} In this stage, we obtain relevance scores for each of the \numdocs documents within the list $\retset_{t}$ we retrieve. As we generate \maxrw{}=5 distinct rewrites for the query, we make relevance assesments for all the \maxrw{} rewrites, resulting in a total of \maxrw{}*\numdocs{}=500 calls to the \gptturbo model. To prevent redundant calls to the model, we maintain a database of previous calls and their corresponding responses. Through the utilization of this database, we observe that our amortized calls for relevance to \gptturbo are significantly reduced, averaging around 300 amortized calls.

\textbf{Rewrite:} We generate a total of \maxrw{}=5 rewrites, with the initial query counting as one, resulting in 4 calls to \gptfour.

\textbf{Rerank:} The re-ranking process involves two steps: (i) Re-ranking all \numdocs documents using \gptturbo, and (ii)Re-ranking the top 30 documents, which were initially re-ranked by \gptturbo, using \gptfour.
This leads to ((\numdocs - \winsize) // \stepsize)+1 calls to \gptturbo and ((30 - \winsize) // \stepsize)+1 calls to \gptfour. Given \winsize{}=10 and \stepsize{}=5, this results in 20 calls to \gptturbo and 5 calls to \gptfour.

Summing up, we make 320 calls to \gptturbo and 9 calls to \gptfour per query.

OpenAI charges for each model are listed below
\begin{itemize}
    \item \gptfour: \$0.03 per 1000 tokens of input and \$0.06 per 1000 tokens of output
    \item \gptturbo: \$0.0015 per 1000 tokens of input and \$0.002 per 1000 tokens of output
\end{itemize}

In our experiments, we notice that the number of output tokens is negligible compared to the number of input tokens. Specifically, when making calls to \gptturbo, approximately 1000 tokens are utilized for input, while calls to \gptfour use around 2000 tokens for input. This results in an approximate cost of \$1.02 per API query, with specific cost details provided in Table \ref{tab:model_calls}.


\begin{table}[!htb]
\caption{The cost of API calls per query.}
\label{tab:model_calls}
\centering
\footnotesize
\begin{tabular}{c | c | c}
    \toprule
    Model & Number of API calls & Cost (\$) \\
    \toprule
    \gptturbo & 320 & 0.48 \\
    \gptfour & 9 & 0.54 \\
    \toprule
\end{tabular}
\end{table}

\section{Hyperparameter Tuning}\label{app:hparam_tuning}


In our final re-ranking step with \gptfour, we conduct hyperparameter tuning for the window size (\winsize) and step size (\stepsize), exploring two specific configurations: \winsize,\stepsize = 20,10 and \winsize,\stepsize = 10,5. Our results indicate that the \winsize,\stepsize = 10,5 configuration outperforms the \winsize,\stepsize = 20,10 setup, with a higher \ndcg{}@10 score of 0.7390 compared to 0.7376.



\section{Additional Results}\label{app:results}
Extended results on \sparta \citep{sparta}, \doctquery \citep{doctquery}, \genq \citep{beir}, \colbert \citep{colbert}, \bmce \citep{bmce}, and \tart \citep{tart} can be found in Table \ref{tab:ndcg_beir_extended} and Table \ref{tab:recall_beir_extended}.

\begin{table*}[!htb]
\caption{Retrieval performance (\ndcg{}@10) on \beir{} datasets.}
\label{tab:ndcg_beir_extended}
\resizebox{1\textwidth}{!}{%
\centering
\begin{tabular}{c c c c c c c c c c}
    \toprule
    Method & \treccovid & \nfcorpus & \signal & \trecnews & \robust & \touche & \dbpedia & \scifact \\ 
    \toprule
    \bm & 59.5 & 30.8 & {33.0} & 39.5 & 40.7 & \textbf{44.2} & 31.8 & 67.9 \\ 
    \sparta & 53.8 & 30.1 & 25.2 & 25.8 & 27.6 & 17.5 & 31.4 & 58.2 \\
    \doctquery & 71.3 & 32.8 & 30.7 & 42.0 & 43.7 & 34.7 & 33.1 & 67.7 \\
    \dpr & 33.2 & 18.9 & 15.5 & 16.1 & 25.2 & 13.1 & 26.3 & 31.8 \\ 
    \ance & 65.4 & 23.7 & 24.9 & 38.2 & 39.2 & 24.0 & 28.1 & 50.7 \\ 
    \tasb & 48.1 & 31.9 & 28.9 & 37.7 & 42.7 & 16.2 & 38.4 & 64.3 \\ 
    \genq & 61.9 & 31.9 & 28.1 & 39.6 & 36.2 & 18.2 & 32.8 & 64.4 \\
    \colbert & 67.7 & 30.5 & 27.4 & 39.3 & 39.1 & 20.2 & 39.2 & 67.1 \\
    \bmce & 75.7 & 35.0 & 33.8 & 43.1 & 27.1 & 40.9 & 68.8 \\
    \monobert (340M) & 70.0 & 36.9 & 31.4 & 44.6 & 49.3 & 31.8 & 41.9 & 71.4 \\
    \monot (220M) & 78.34 & 37.4 & 31.7 & 46.8 & 51.7 & 30.8 & 42.4 & 73.4 \\
    \monot (3B) & 80.7 & {39.0} & 32.6 & 48.5 & 56.7 & 32.4 & 44.4 & {76.6} \\ 
    \tart & 75.1 & 36.0 & 25.8 & 40.0 & 50.8 & 27.5 & 42.5 & 74.8 \\
    \upr & 68.1 & 35.0 & 31.9 & 43.1 & 42.4 & 19.7 & 30.9 & 72.7 \\
    \pgator++(zero-shot) & 76.0 & 36.0 & - & - & - & 27.8 & 41.3 & 73.6 \\
    \pgator++(few-shot) & 76.2 & 37.0 & - & - & - & 38.1 & 43.4 & 73.1 \\ 
    \rankgpt & {85.5} & 38.5 & \textbf{34.4} & {52.9} & {57.6} & {38.6} & {47.1} & 75.0 \\ 
    \rrr (this work) & \textbf{86.4} & \textbf{39.9} & 29.8 & \textbf{53.6} & \textbf{67.4} & 29.8 & \textbf{51.0} & \textbf{77.2} \\ 
    \toprule
\end{tabular}
}
\end{table*}

\begin{table*}[!htb]
\caption{Retrieval performance (\recall{}@100) on \beir{} datasets. For \treccovid, capped \recall{}@100 is used.}
\label{tab:recall_beir_extended}
\resizebox{1\textwidth}{!}{%
\centering
\begin{tabular}{c c c c c c c c c c}
    \toprule
    Method & \treccovid & \nfcorpus & \signal & \trecnews & \robust & \touche & \dbpedia & \scifact \\ 
    \toprule
    \bm  &  {49.8}  & 24.6 &  \textbf{37.0}  &  {44.7}  &  {37.5}  &  \textbf{58.2}  & 46.8 &  {92.5} \\ 
    \sparta & 40.9 & 24.3 & 27.0 & 26.2 & 21.5 & 38.1 & 41.1 & 86.3 \\
    \doctquery & 54.1 & 25.3 & 35.1 & 43.9 & 35.7 & 55.7 & 36.5 & 91.4 \\
    \dpr  & 21.2 & 20.8 & 16.2 & 21.5 & 21.1 & 30.1 & 34.9 & 72.7 \\ 
    \ance  & 45.7 & 23.2 & 23.9 & 39.8 & 27.4 & 45.8 & 31.9 & 81.6 \\ 
    \tasb  & 38.7 &  {28.0}  & 30.4 & 41.8 & 33.1 & 43.1 &  {49.9}  & 89.1 \\ 
    \genq & 45.6 & 28.0 & 28.1 & 41.2 & 29.8 & 45.1 & 43.1 & 89.3 \\
    \colbert & 46.4 & 25.4 & 28.3 & 36.7 & 31.0 & 43.9 & 46.1 & 87.8 \\
    \bmce & {49.8}  & 24.6 &  \textbf{37.0}  &  {44.7}  &  {37.5}  &  \textbf{58.2}  & 46.8 &  {92.5} \\
    \monobert (340M) & {49.8}  & 24.6 &  \textbf{37.0}  &  {44.7}  &  {37.5}  &  \textbf{58.2}  & 46.8 &  {92.5} \\
    \monot (220M) & {49.8}  & 24.6 &  \textbf{37.0}  &  {44.7}  &  {37.5}  &  \textbf{58.2}  & 46.8 &  {92.5} \\
    \monot (3B)  &  {49.8}  & 24.6 &  \textbf{37.0}  &  {44.7}  &  {37.5}  &  \textbf{58.2}  & 46.8 &  {92.5} \\ 
    \rankgpt  &  {49.8}  & 24.6 &  \textbf{37.0}  &  {44.7}  &  {37.5}  &  \textbf{58.2}  & 46.8 &  {92.5} \\ 
    \rrr (this work)  &  \textbf{54.8}  &  \textbf{32.4}  &  {32.4}  &  \textbf{51.6}  &  \textbf{45.4}   &  {52.2}  &  \textbf{55.0}  &  \textbf{94.3} \\ 
    \toprule
\end{tabular}
}
\end{table*}


%% file: main.bbl
\begin{thebibliography}{71}
\providecommand{\natexlab}[1]{#1}
\providecommand{\url}[1]{\texttt{#1}}
\expandafter\ifx\csname urlstyle\endcsname\relax
  \providecommand{\doi}[1]{doi: #1}\else
  \providecommand{\doi}{doi: \begingroup \urlstyle{rm}\Url}\fi

\bibitem[Abdul-Jaleel et~al.(2004)Abdul-Jaleel, Allan, Croft, Diaz, Larkey, Li, Smucker, and Wade]{abdul2004umass}
Nasreen Abdul-Jaleel, James Allan, W~Bruce Croft, Fernando Diaz, Leah Larkey, Xiaoyan Li, Mark~D Smucker, and Courtney Wade.
\newblock Umass at trec 2004: Novelty and hard.
\newblock \emph{Computer Science Department Faculty Publication Series}, pp.\  189, 2004.

\bibitem[Asai et~al.(2022)Asai, Schick, Lewis, Chen, Izacard, Riedel, Hajishirzi, and tau Yih]{tart}
Akari Asai, Timo Schick, Patrick Lewis, Xilun Chen, Gautier Izacard, Sebastian Riedel, Hannaneh Hajishirzi, and Wen tau Yih.
\newblock Task-aware retrieval with instructions, 2022.
\newblock URL \url{https://arxiv.org/abs/2211.09260}.

\bibitem[Bai et~al.(2022)Bai, Kadavath, Kundu, Askell, Kernion, Jones, Chen, Goldie, Mirhoseini, McKinnon, et~al.]{bai2022constitutional}
Yuntao Bai, Saurav Kadavath, Sandipan Kundu, Amanda Askell, Jackson Kernion, Andy Jones, Anna Chen, Anna Goldie, Azalia Mirhoseini, Cameron McKinnon, et~al.
\newblock Constitutional ai: Harmlessness from ai feedback.
\newblock \emph{arXiv preprint arXiv:2212.08073}, 2022.
\newblock URL \url{https://arxiv.org/abs/2212.08073}.

\bibitem[Bajaj et~al.(2016)Bajaj, Campos, Craswell, Deng, Gao, Liu, Majumder, McNamara, Mitra, Nguyen, et~al.]{nguyen2640ms}
Payal Bajaj, Daniel Campos, Nick Craswell, Li~Deng, Jianfeng Gao, Xiaodong Liu, Rangan Majumder, Andrew McNamara, Bhaskar Mitra, Tri Nguyen, et~al.
\newblock Ms marco: A human generated machine reading comprehension dataset.
\newblock \emph{arXiv preprint arXiv:1611.09268}, 2016.
\newblock URL \url{https://arxiv.org/abs/1611.09268}.

\bibitem[Belkin et~al.(2003)Belkin, Kelly, Kim, Kim, Lee, Muresan, Tang, Yuan, and Cool]{belkin03search}
N.~J. Belkin, D.~Kelly, G.~Kim, J.-Y. Kim, H.-J. Lee, G.~Muresan, M.-C. Tang, X.-J. Yuan, and C.~Cool.
\newblock Query length in interactive information retrieval.
\newblock In \emph{Proceedings of the 26th Annual International ACM SIGIR Conference on Research and Development in Informaion Retrieval}, SIGIR '03, pp.\  205–212, New York, NY, USA, 2003. Association for Computing Machinery.
\newblock ISBN 1581136463.
\newblock \doi{10.1145/860435.860474}.
\newblock URL \url{https://doi.org/10.1145/860435.860474}.

\bibitem[Bonifacio et~al.(2022)Bonifacio, Abonizio, Fadaee, and Nogueira]{inparsv1}
Luiz Bonifacio, Hugo Abonizio, Marzieh Fadaee, and Rodrigo Nogueira.
\newblock {InPars}: Unsupervised dataset generation for information retrieval.
\newblock In \emph{Proceedings of the 45th International ACM SIGIR Conference on Research and Development in Information Retrieval}, SIGIR '22, pp.\  2387–2392, New York, NY, USA, 2022. Association for Computing Machinery.
\newblock ISBN 9781450387323.
\newblock \doi{10.1145/3477495.3531863}.
\newblock URL \url{https://doi.org/10.1145/3477495.3531863}.

\bibitem[Bonifacio et~al.(2021)Bonifacio, Campiotti, de~Alencar~Lotufo, and Nogueira]{mmarco}
Luiz~Henrique Bonifacio, Israel Campiotti, Roberto de~Alencar~Lotufo, and Rodrigo~Frassetto Nogueira.
\newblock mmarco: {A} multilingual version of {MS} {MARCO} passage ranking dataset.
\newblock \emph{CoRR}, abs/2108.13897, 2021.
\newblock URL \url{https://arxiv.org/abs/2108.13897}.

\bibitem[Brill et~al.(2002)Brill, Dumais, and Banko]{brill2002analysis}
Eric Brill, Susan Dumais, and Michele Banko.
\newblock An analysis of the {A}sk{MSR} question-answering system.
\newblock In \emph{Proceedings of the 2002 Conference on Empirical Methods in Natural Language Processing ({EMNLP} 2002)}, pp.\  257--264. Association for Computational Linguistics, July 2002.
\newblock \doi{10.3115/1118693.1118726}.
\newblock URL \url{https://aclanthology.org/W02-1033}.

\bibitem[Chen et~al.(2017)Chen, Fisch, Weston, and Bordes]{chen2017reading}
Danqi Chen, Adam Fisch, Jason Weston, and Antoine Bordes.
\newblock Reading wikipedia to answer open-domain questions.
\newblock In \emph{55th Annual Meeting of the Association for Computational Linguistics, ACL 2017}, pp.\  1870--1879. Association for Computational Linguistics (ACL), 2017.

\bibitem[Cheriton(2019)]{doctquery}
David~R. Cheriton.
\newblock From doc2query to doctttttquery.
\newblock 2019.
\newblock URL \url{https://api.semanticscholar.org/CorpusID:208612557}.

\bibitem[Covington et~al.(2016)Covington, Adams, and Sargin]{covington16youtube}
Paul Covington, Jay Adams, and Emre Sargin.
\newblock Deep neural networks for youtube recommendations.
\newblock In \emph{Proceedings of the 10th ACM Conference on Recommender Systems}, RecSys '16, pp.\  191–198, New York, NY, USA, 2016. Association for Computing Machinery.
\newblock ISBN 9781450340359.
\newblock \doi{10.1145/2959100.2959190}.
\newblock URL \url{https://doi.org/10.1145/2959100.2959190}.

\bibitem[Craswell et~al.(2020{\natexlab{a}})Craswell, Mitra, Yilmaz, Campos, and Voorhees]{trecdl19}
Nick Craswell, Bhaskar Mitra, Emine Yilmaz, Daniel Campos, and Ellen~M. Voorhees.
\newblock Overview of the {TREC} 2019 deep learning track.
\newblock \emph{CoRR}, abs/2003.07820, 2020{\natexlab{a}}.
\newblock URL \url{https://arxiv.org/abs/2003.07820}.

\bibitem[Craswell et~al.(2020{\natexlab{b}})Craswell, Mitra, Yilmaz, Campos, and Voorhees]{trecdl20}
Nick Craswell, Bhaskar Mitra, Emine Yilmaz, Daniel Campos, and Ellen~M. Voorhees.
\newblock Overview of the {TREC} 2019 deep learning track.
\newblock \emph{CoRR}, abs/2003.07820, 2020{\natexlab{b}}.
\newblock URL \url{https://arxiv.org/abs/2003.07820}.

\bibitem[Dahiya et~al.(2021)Dahiya, Saini, Mittal, Shaw, Dave, Soni, Jain, Agarwal, and Varma]{dahiya21xml}
Kunal Dahiya, Deepak Saini, Anshul Mittal, Ankush Shaw, Kushal Dave, Akshay Soni, Himanshu Jain, Sumeet Agarwal, and Manik Varma.
\newblock {DeepXML}: A deep extreme multi-label learning framework applied to short text documents.
\newblock In \emph{Proceedings of the 14th ACM International Conference on Web Search and Data Mining}, WSDM '21, pp.\  31–39, New York, NY, USA, 2021. Association for Computing Machinery.
\newblock ISBN 9781450382977.
\newblock \doi{10.1145/3437963.3441810}.
\newblock URL \url{https://doi.org/10.1145/3437963.3441810}.

\bibitem[Dai et~al.(2022)Dai, Zhao, Ma, Luan, Ni, Lu, Bakalov, Guu, Hall, and Chang]{pgator22}
Zhuyun Dai, Vincent~Y. Zhao, Ji~Ma, Yi~Luan, Jianmo Ni, Jing Lu, Anton Bakalov, Kelvin Guu, Keith~B. Hall, and Ming-Wei Chang.
\newblock Promptagator: Few-shot dense retrieval from 8 examples, 2022.
\newblock URL \url{https://arxiv.org/abs/2209.11755}.

\bibitem[Feng et~al.(2023)Feng, Tao, Geng, Shen, Xu, Long, Zhao, and Jiang]{feng23knowledge}
Jiazhan Feng, Chongyang Tao, Xiubo Geng, Tao Shen, Can Xu, Guodong Long, Dongyan Zhao, and Daxin Jiang.
\newblock Knowledge refinement via interaction between search engines and large language models, 2023.
\newblock URL \url{https://arxiv.org/abs/2305.07402}.

\bibitem[Gao et~al.(2022)Gao, Ma, Lin, and Callan]{hyde22}
Luyu Gao, Xueguang Ma, Jimmy Lin, and Jamie Callan.
\newblock Precise zero-shot dense retrieval without relevance labels, 2022.
\newblock URL \url{https://arxiv.org/abs/2212.10496}.

\bibitem[Guu et~al.(2020)Guu, Lee, Tung, Pasupat, and Chang]{guu2020retrieval}
Kelvin Guu, Kenton Lee, Zora Tung, Panupong Pasupat, and Mingwei Chang.
\newblock Retrieval augmented language model pre-training.
\newblock In \emph{International conference on machine learning}, pp.\  3929--3938. PMLR, 2020.

\bibitem[Hofstätter et~al.(2021)Hofstätter, Lin, Yang, Lin, and Hanbury]{tasb}
Sebastian Hofstätter, Sheng-Chieh Lin, Jheng-Hong Yang, Jimmy Lin, and Allan Hanbury.
\newblock Efficiently teaching an effective dense retriever with balanced topic aware sampling, 2021.
\newblock URL \url{https://arxiv.org/abs/2104.06967}.

\bibitem[Izacard et~al.(2022)Izacard, Lewis, Lomeli, Hosseini, Petroni, Schick, Dwivedi-Yu, Joulin, Riedel, and Grave]{izacard2022few}
Gautier Izacard, Patrick Lewis, Maria Lomeli, Lucas Hosseini, Fabio Petroni, Timo Schick, Jane Dwivedi-Yu, Armand Joulin, Sebastian Riedel, and Edouard Grave.
\newblock Few-shot learning with retrieval augmented language models.
\newblock \emph{arXiv preprint arXiv:2208.03299}, 2022.
\newblock URL \url{https://arxiv.org/abs/2208.03299}.

\bibitem[Jagerman et~al.(2023)Jagerman, Zhuang, Qin, Wang, and Bendersky]{jagerman23}
Rolf Jagerman, Honglei Zhuang, Zhen Qin, Xuanhui Wang, and Michael Bendersky.
\newblock Query expansion by prompting large language models, 2023.
\newblock URL \url{https://arxiv.org/abs/2305.03653}.

\bibitem[J{\"a}rvelin \& Kek{\"a}l{\"a}inen(2002)J{\"a}rvelin and Kek{\"a}l{\"a}inen]{jarvelin2002cumulated}
Kalervo J{\"a}rvelin and Jaana Kek{\"a}l{\"a}inen.
\newblock Cumulated gain-based evaluation of {IR} techniques.
\newblock \emph{ACM Transactions on Information Systems (TOIS)}, 20\penalty0 (4):\penalty0 422--446, 2002.

\bibitem[Jeronymo et~al.(2023)Jeronymo, Bonifacio, Abonizio, Fadaee, Lotufo, Zavrel, and Nogueira]{inparsv2}
Vitor Jeronymo, Luiz Bonifacio, Hugo Abonizio, Marzieh Fadaee, Roberto Lotufo, Jakub Zavrel, and Rodrigo Nogueira.
\newblock {InPars-v2}: Large language models as efficient dataset generators for information retrieval, 2023.
\newblock URL \url{https://arxiv.org/abs/2301.01820}.

\bibitem[Karpukhin et~al.(2020)Karpukhin, Oguz, Min, Lewis, Wu, Edunov, Chen, and Yih]{dpr}
Vladimir Karpukhin, Barlas Oguz, Sewon Min, Patrick Lewis, Ledell Wu, Sergey Edunov, Danqi Chen, and Wen-tau Yih.
\newblock Dense passage retrieval for open-domain question answering.
\newblock In \emph{Proceedings of the 2020 Conference on Empirical Methods in Natural Language Processing (EMNLP)}, pp.\  6769--6781, Online, November 2020. Association for Computational Linguistics.
\newblock \doi{10.18653/v1/2020.emnlp-main.550}.
\newblock URL \url{https://aclanthology.org/2020.emnlp-main.550}.

\bibitem[Khattab \& Zaharia(2020)Khattab and Zaharia]{colbert}
Omar Khattab and Matei Zaharia.
\newblock Colbert: Efficient and effective passage search via contextualized late interaction over bert, 2020.
\newblock URL \url{https://arxiv.org/abs/2004.12832}.

\bibitem[Kobayashi \& Takeda(2000)Kobayashi and Takeda]{kobayashi2000information}
Mei Kobayashi and Koichi Takeda.
\newblock Information retrieval on the web.
\newblock \emph{ACM computing surveys (CSUR)}, 32\penalty0 (2):\penalty0 144--173, 2000.

\bibitem[Kocmi \& Federmann(2023)Kocmi and Federmann]{kocmi2023large}
Tom Kocmi and Christian Federmann.
\newblock Large language models are state-of-the-art evaluators of translation quality.
\newblock \emph{arXiv preprint arXiv:2302.14520}, 2023.
\newblock URL \url{https://arxiv.org/abs/2302.14520}.

\bibitem[Lee et~al.(2019)Lee, Chang, and Toutanova]{lee2019latent}
Kenton Lee, Ming-Wei Chang, and Kristina Toutanova.
\newblock Latent retrieval for weakly supervised open domain question answering.
\newblock In \emph{Proceedings of the 57th Annual Meeting of the Association for Computational Linguistics}, pp.\  6086--6096, Florence, Italy, July 2019. Association for Computational Linguistics.
\newblock \doi{10.18653/v1/P19-1612}.
\newblock URL \url{https://aclanthology.org/P19-1612}.

\bibitem[Lewis et~al.(2020)Lewis, Perez, Piktus, Petroni, Karpukhin, Goyal, K\"{u}ttler, Lewis, Yih, Rockt\"{a}schel, Riedel, and Kiela]{lewis2020retrieval}
Patrick Lewis, Ethan Perez, Aleksandra Piktus, Fabio Petroni, Vladimir Karpukhin, Naman Goyal, Heinrich K\"{u}ttler, Mike Lewis, Wen-tau Yih, Tim Rockt\"{a}schel, Sebastian Riedel, and Douwe Kiela.
\newblock Retrieval-augmented generation for knowledge-intensive nlp tasks.
\newblock In H.~Larochelle, M.~Ranzato, R.~Hadsell, M.F. Balcan, and H.~Lin (eds.), \emph{Advances in Neural Information Processing Systems}, volume~33, pp.\  9459--9474. Curran Associates, Inc., 2020.

\bibitem[Liang et~al.(2023)Liang, Bommasani, Lee, Tsipras, Soylu, Yasunaga, Zhang, Narayanan, Wu, Kumar, Newman, Yuan, Yan, Zhang, Cosgrove, Manning, Ré, Acosta-Navas, Hudson, Zelikman, Durmus, Ladhak, Rong, Ren, Yao, Wang, Santhanam, Orr, Zheng, Yuksekgonul, Suzgun, Kim, Guha, Chatterji, Khattab, Henderson, Huang, Chi, Xie, Santurkar, Ganguli, Hashimoto, Icard, Zhang, Chaudhary, Wang, Li, Mai, Zhang, and Koreeda]{liang2023holistic}
Percy Liang, Rishi Bommasani, Tony Lee, Dimitris Tsipras, Dilara Soylu, Michihiro Yasunaga, Yian Zhang, Deepak Narayanan, Yuhuai Wu, Ananya Kumar, Benjamin Newman, Binhang Yuan, Bobby Yan, Ce~Zhang, Christian Cosgrove, Christopher~D. Manning, Christopher Ré, Diana Acosta-Navas, Drew~A. Hudson, Eric Zelikman, Esin Durmus, Faisal Ladhak, Frieda Rong, Hongyu Ren, Huaxiu Yao, Jue Wang, Keshav Santhanam, Laurel Orr, Lucia Zheng, Mert Yuksekgonul, Mirac Suzgun, Nathan Kim, Neel Guha, Niladri Chatterji, Omar Khattab, Peter Henderson, Qian Huang, Ryan Chi, Sang~Michael Xie, Shibani Santurkar, Surya Ganguli, Tatsunori Hashimoto, Thomas Icard, Tianyi Zhang, Vishrav Chaudhary, William Wang, Xuechen Li, Yifan Mai, Yuhui Zhang, and Yuta Koreeda.
\newblock Holistic evaluation of language models, 2023.
\newblock URL \url{https://arxiv.org/abs/2211.09110}.

\bibitem[Lin et~al.(2021)Lin, Ma, Lin, Yang, Pradeep, and Nogueira]{pyserini}
Jimmy Lin, Xueguang Ma, Sheng-Chieh Lin, Jheng-Hong Yang, Ronak Pradeep, and Rodrigo Nogueira.
\newblock {Pyserini}: A {Python} toolkit for reproducible information retrieval research with sparse and dense representations.
\newblock In \emph{Proceedings of the 44th Annual International ACM SIGIR Conference on Research and Development in Information Retrieval (SIGIR 2021)}, pp.\  2356--2362, 2021.

\bibitem[Lin et~al.(2023)Lin, Asai, Li, Oguz, Lin, Mehdad, tau Yih, and Chen]{dragon23}
Sheng-Chieh Lin, Akari Asai, Minghan Li, Barlas Oguz, Jimmy Lin, Yashar Mehdad, Wen tau Yih, and Xilun Chen.
\newblock How to train your dragon: Diverse augmentation towards generalizable dense retrieval, 2023.
\newblock URL \url{https://arxiv.org/abs/2302.07452}.

\bibitem[Mackie et~al.(2023{\natexlab{a}})Mackie, Chatterjee, and Dalton]{generativeprf23}
Iain Mackie, Shubham Chatterjee, and Jeffrey Dalton.
\newblock Generative and pseudo-relevant feedback for sparse, dense and learned sparse retrieval, 2023{\natexlab{a}}.
\newblock URL \url{https://arxiv.org/abs/2305.07477}.

\bibitem[Mackie et~al.(2023{\natexlab{b}})Mackie, Chatterjee, and Dalton]{genfeedback23}
Iain Mackie, Shubham Chatterjee, and Jeffrey Dalton.
\newblock Generative relevance feedback with large language models.
\newblock In \emph{Proceedings of the 46th International ACM SIGIR Conference on Research and Development in Information Retrieval}, SIGIR '23, pp.\  2026–2031, New York, NY, USA, 2023{\natexlab{b}}. Association for Computing Machinery.
\newblock ISBN 9781450394086.
\newblock \doi{10.1145/3539618.3591992}.
\newblock URL \url{https://doi.org/10.1145/3539618.3591992}.

\bibitem[Mackie et~al.(2023{\natexlab{c}})Mackie, Sekulic, Chatterjee, Dalton, and Crestani]{grm23}
Iain Mackie, Ivan Sekulic, Shubham Chatterjee, Jeffrey Dalton, and Fabio Crestani.
\newblock Grm: Generative relevance modeling using relevance-aware sample estimation for document retrieval, 2023{\natexlab{c}}.
\newblock URL \url{https://arxiv.org/abs/2306.09938}.

\bibitem[Mao et~al.(2023)Mao, Dou, Chen, Mo, and Qian]{conversationsearch23}
Kelong Mao, Zhicheng Dou, Haonan Chen, Fengran Mo, and Hongjin Qian.
\newblock Large language models know your contextual search intent: A prompting framework for conversational search, 2023.
\newblock URL \url{https://arxiv.org/abs/2303.06573}.

\bibitem[Nogueira et~al.(2019{\natexlab{a}})Nogueira, Yang, Cho, and Lin]{nogueira2019multi}
Rodrigo Nogueira, Wei Yang, Kyunghyun Cho, and Jimmy Lin.
\newblock Multi-stage document ranking with bert.
\newblock \emph{arXiv e-prints}, pp.\  arXiv--1910, 2019{\natexlab{a}}.
\newblock URL \url{https://arxiv.org/abs/1910.14424}.

\bibitem[Nogueira et~al.(2019{\natexlab{b}})Nogueira, Yang, Lin, and Cho]{nogueira2019document}
Rodrigo Nogueira, Wei Yang, Jimmy Lin, and Kyunghyun Cho.
\newblock Document expansion by query prediction, 2019{\natexlab{b}}.
\newblock URL \url{https://arxiv.org/abs/1904.08375}.

\bibitem[Nogueira et~al.(2020)Nogueira, Jiang, Pradeep, and Lin]{monot}
Rodrigo Nogueira, Zhiying Jiang, Ronak Pradeep, and Jimmy Lin.
\newblock Document ranking with a pretrained sequence-to-sequence model.
\newblock In \emph{Findings of the Association for Computational Linguistics: EMNLP 2020}, pp.\  708--718, Online, November 2020. Association for Computational Linguistics.
\newblock \doi{10.18653/v1/2020.findings-emnlp.63}.
\newblock URL \url{https://aclanthology.org/2020.findings-emnlp.63}.

\bibitem[Nogueira \& Cho(2019)Nogueira and Cho]{monobert}
Rodrigo~Frassetto Nogueira and Kyunghyun Cho.
\newblock Passage re-ranking with {BERT}.
\newblock \emph{CoRR}, abs/1901.04085, 2019.
\newblock URL \url{http://arxiv.org/abs/1901.04085}.

\bibitem[OpenAI(2022)]{chatgpt}
OpenAI.
\newblock {Introducing ChatGPT}.
\newblock https://openai.com/blog/chatgpt, 2022.
\newblock Accessed: 30th September, 2023.

\bibitem[OpenAI(2023)]{gpt4}
OpenAI.
\newblock {GPT-4 Technical Report}, 2023.
\newblock URL \url{https://arxiv.org/abs/2303.08774}.

\bibitem[Parvez et~al.(2021)Parvez, Ahmad, Chakraborty, Ray, and Chang]{parvez2021retrieval}
Md~Rizwan Parvez, Wasi Ahmad, Saikat Chakraborty, Baishakhi Ray, and Kai-Wei Chang.
\newblock Retrieval augmented code generation and summarization.
\newblock In \emph{Findings of the Association for Computational Linguistics: EMNLP 2021}, pp.\  2719--2734, Punta Cana, Dominican Republic, November 2021. Association for Computational Linguistics.
\newblock \doi{10.18653/v1/2021.findings-emnlp.232}.
\newblock URL \url{https://aclanthology.org/2021.findings-emnlp.232}.

\bibitem[Pinkerton(1994)]{pinkerton1994finding}
Brian Pinkerton.
\newblock Finding what people want: Experiences with the webcrawler.
\newblock In \emph{Proc. of the 2nd Int. World Wide Web Conf., 1994}, 1994.

\bibitem[Qin et~al.(2023)Qin, Jagerman, Hui, Zhuang, Wu, Shen, Liu, Liu, Metzler, Wang, and Bendersky]{qinllm23}
Zhen Qin, Rolf Jagerman, Kai Hui, Honglei Zhuang, Junru Wu, Jiaming Shen, Tianqi Liu, Jialu Liu, Donald Metzler, Xuanhui Wang, and Michael Bendersky.
\newblock Large language models are effective text rankers with pairwise ranking prompting, 2023.
\newblock URL \url{https://arxiv.org/abs/2306.17563}.

\bibitem[Qu et~al.(2021)Qu, Ding, Liu, Liu, Ren, Zhao, Dong, Wu, and Wang]{qu2021rocketqa}
Yingqi Qu, Yuchen Ding, Jing Liu, Kai Liu, Ruiyang Ren, Wayne~Xin Zhao, Daxiang Dong, Hua Wu, and Haifeng Wang.
\newblock Rocketqa: An optimized training approach to dense passage retrieval for open-domain question answering.
\newblock In \emph{Proceedings of the 2021 Conference of the North American Chapter of the Association for Computational Linguistics: Human Language Technologies}, pp.\  5835--5847, 2021.
\newblock URL \url{https://aclanthology.org/2021.naacl-main.466}.

\bibitem[Roberts et~al.(2020)Roberts, Raffel, and Shazeer]{roberts-etal-2020-much}
Adam Roberts, Colin Raffel, and Noam Shazeer.
\newblock How much knowledge can you pack into the parameters of a language model?
\newblock In \emph{Proceedings of the 2020 Conference on Empirical Methods in Natural Language Processing (EMNLP)}, pp.\  5418--5426, Online, November 2020. Association for Computational Linguistics.
\newblock \doi{10.18653/v1/2020.emnlp-main.437}.
\newblock URL \url{https://aclanthology.org/2020.emnlp-main.437}.

\bibitem[Robertson \& Zaragoza(2009)Robertson and Zaragoza]{bm25}
Stephen Robertson and Hugo Zaragoza.
\newblock The probabilistic relevance framework: {BM25} and {B}eyond.
\newblock \emph{Foundations and Trends® in Information Retrieval}, 3\penalty0 (4):\penalty0 333--389, 2009.
\newblock ISSN 1554-0669.
\newblock \doi{10.1561/1500000019}.
\newblock URL \url{http://dx.doi.org/10.1561/1500000019}.

\bibitem[Ruthven(2008)]{Ian08ir}
Ian Ruthven.
\newblock Interactive information retrieval.
\newblock \emph{Annual Review of Information Science and Technology}, 42\penalty0 (1):\penalty0 43--91, 2008.
\newblock \doi{https://doi.org/10.1002/aris.2008.1440420109}.
\newblock URL \url{https://asistdl.onlinelibrary.wiley.com/doi/abs/10.1002/aris.2008.1440420109}.

\bibitem[Sachan et~al.(2022)Sachan, Lewis, Joshi, Aghajanyan, Yih, Pineau, and Zettlemoyer]{upr}
Devendra Sachan, Mike Lewis, Mandar Joshi, Armen Aghajanyan, Wen-tau Yih, Joelle Pineau, and Luke Zettlemoyer.
\newblock Improving passage retrieval with zero-shot question generation.
\newblock In \emph{Proceedings of the 2022 Conference on Empirical Methods in Natural Language Processing}, pp.\  3781--3797, Abu Dhabi, United Arab Emirates, December 2022. Association for Computational Linguistics.
\newblock \doi{10.18653/v1/2022.emnlp-main.249}.
\newblock URL \url{https://aclanthology.org/2022.emnlp-main.249}.

\bibitem[Sachan et~al.(2023)Sachan, Lewis, Yogatama, Zettlemoyer, Pineau, and Zaheer]{sachan23art}
Devendra~Singh Sachan, Mike Lewis, Dani Yogatama, Luke Zettlemoyer, Joelle Pineau, and Manzil Zaheer.
\newblock {Questions Are All You Need to Train a Dense Passage Retriever}.
\newblock \emph{Transactions of the Association for Computational Linguistics}, 11:\penalty0 600--616, 06 2023.
\newblock ISSN 2307-387X.
\newblock \doi{10.1162/tacl_a_00564}.
\newblock URL \url{https://doi.org/10.1162/tacl\_a\_00564}.

\bibitem[Salton \& Buckley(1988)Salton and Buckley]{salton1988term}
Gerard Salton and Christopher Buckley.
\newblock Term-weighting approaches in automatic text retrieval.
\newblock \emph{Information processing \& management}, 24\penalty0 (5):\penalty0 513--523, 1988.

\bibitem[Shen et~al.(2023)Shen, Long, Geng, Tao, Zhou, and Jiang]{shen23}
Tao Shen, Guodong Long, Xiubo Geng, Chongyang Tao, Tianyi Zhou, and Daxin Jiang.
\newblock Large language models are strong zero-shot retriever, 2023.
\newblock URL \url{https://arxiv.org/abs/2304.14233}.

\bibitem[Singh et~al.(2021)Singh, Reddy, Hamilton, Dyer, and Yogatama]{singh2021end}
Devendra Singh, Siva Reddy, Will Hamilton, Chris Dyer, and Dani Yogatama.
\newblock End-to-end training of multi-document reader and retriever for open-domain question answering.
\newblock In M.~Ranzato, A.~Beygelzimer, Y.~Dauphin, P.S. Liang, and J.~Wortman Vaughan (eds.), \emph{Advances in Neural Information Processing Systems}, volume~34, pp.\  25968--25981. Curran Associates, Inc., 2021.

\bibitem[Srinivasan et~al.(2022)Srinivasan, Raman, Samanta, Liao, Bertelli, and Bendersky]{quill22}
Krishna Srinivasan, Karthik Raman, Anupam Samanta, Lingrui Liao, Luca Bertelli, and Michael Bendersky.
\newblock {QUILL}: Query intent with large language models using retrieval augmentation and multi-stage distillation.
\newblock In \emph{Proceedings of the 2022 Conference on Empirical Methods in Natural Language Processing: Industry Track}, pp.\  492--501, Abu Dhabi, UAE, December 2022. Association for Computational Linguistics.
\newblock \doi{10.18653/v1/2022.emnlp-industry.50}.
\newblock URL \url{https://aclanthology.org/2022.emnlp-industry.50}.

\bibitem[Su \& Khoshgoftaar(2009)Su and Khoshgoftaar]{su09reco}
Xiaoyuan Su and Taghi~M. Khoshgoftaar.
\newblock A survey of collaborative filtering techniques.
\newblock \emph{Adv. in Artif. Intell.}, 2009, jan 2009.
\newblock ISSN 1687-7470.
\newblock \doi{10.1155/2009/421425}.
\newblock URL \url{https://doi.org/10.1155/2009/421425}.

\bibitem[Sun et~al.(2023)Sun, Yan, Ma, Ren, Yin, and Ren]{sungpt23}
Weiwei Sun, Lingyong Yan, Xinyu Ma, Pengjie Ren, Dawei Yin, and Zhaochun Ren.
\newblock Is {ChatGPT} good at search? investigating large language models as re-ranking agent.
\newblock In \emph{Appearing in EMNLP}, 2023.
\newblock URL \url{https://arxiv.org/abs/2304.09542}.

\bibitem[Thakur et~al.(2021)Thakur, Reimers, R{\"{u}}ckl{\'{e}}, Srivastava, and Gurevych]{beir}
Nandan Thakur, Nils Reimers, Andreas R{\"{u}}ckl{\'{e}}, Abhishek Srivastava, and Iryna Gurevych.
\newblock {BEIR:} {A} heterogenous benchmark for zero-shot evaluation of information retrieval models.
\newblock \emph{CoRR}, abs/2104.08663, 2021.
\newblock URL \url{https://arxiv.org/abs/2104.08663}.

\bibitem[Vemuri et~al.(2023)Vemuri, Agrawal, Mittal, Saini, Soni, Sambasivan, Lu, Wang, Parsana, Kar, and Varma]{vemuri23xc}
Hemanth Vemuri, Sheshansh Agrawal, Shivam Mittal, Deepak Saini, Akshay Soni, Abhinav~V. Sambasivan, Wenhao Lu, Yajun Wang, Mehul Parsana, Purushottam Kar, and Manik Varma.
\newblock Personalized retrieval over millions of items.
\newblock In \emph{Proceedings of the 46th International ACM SIGIR Conference on Research and Development in Information Retrieval}, SIGIR '23, pp.\  1014–1022, New York, NY, USA, 2023. Association for Computing Machinery.
\newblock ISBN 9781450394086.
\newblock \doi{10.1145/3539618.3591749}.
\newblock URL \url{https://doi.org/10.1145/3539618.3591749}.

\bibitem[Wang et~al.(2023)Wang, Yang, and Wei]{query2doc23}
Liang Wang, Nan Yang, and Furu Wei.
\newblock Query2doc: Query expansion with large language models, 2023.
\newblock URL \url{https://arxiv.org/abs/2303.07678}.

\bibitem[Wang et~al.(2020)Wang, Wei, Dong, Bao, Yang, and Zhou]{bmce}
Wenhui Wang, Furu Wei, Li~Dong, Hangbo Bao, Nan Yang, and Ming Zhou.
\newblock Minilm: Deep self-attention distillation for task-agnostic compression of pre-trained transformers.
\newblock In H.~Larochelle, M.~Ranzato, R.~Hadsell, M.F. Balcan, and H.~Lin (eds.), \emph{Advances in Neural Information Processing Systems}, volume~33, pp.\  5776--5788. Curran Associates, Inc., 2020.

\bibitem[Wang et~al.(2021)Wang, Macdonald, Tonellotto, and Ounis]{wang2021pseudo}
Xiao Wang, Craig Macdonald, Nicola Tonellotto, and Iadh Ounis.
\newblock Pseudo-relevance feedback for multiple representation dense retrieval.
\newblock In \emph{Proceedings of the 2021 ACM SIGIR International Conference on Theory of Information Retrieval}, ICTIR '21, pp.\  297–306, New York, NY, USA, 2021. Association for Computing Machinery.
\newblock ISBN 9781450386111.
\newblock \doi{10.1145/3471158.3472250}.
\newblock URL \url{https://doi.org/10.1145/3471158.3472250}.

\bibitem[Xiong et~al.(2020)Xiong, Xiong, Li, Tang, Liu, Bennett, Ahmed, and Overwijk]{ance}
Lee Xiong, Chenyan Xiong, Ye~Li, Kwok-Fung Tang, Jialin Liu, Paul~N Bennett, Junaid Ahmed, and Arnold Overwijk.
\newblock Approximate nearest neighbor negative contrastive learning for dense text retrieval.
\newblock In \emph{International Conference on Learning Representations}, 2020.

\bibitem[Yang et~al.(2019)Yang, Xie, Lin, Li, Tan, Xiong, Li, and Lin]{yang-etal-2019-end-end}
Wei Yang, Yuqing Xie, Aileen Lin, Xingyu Li, Luchen Tan, Kun Xiong, Ming Li, and Jimmy Lin.
\newblock End-to-end open-domain question answering with {BERT}serini.
\newblock In \emph{Proceedings of the 2019 Conference of the North {A}merican Chapter of the Association for Computational Linguistics (Demonstrations)}, pp.\  72--77, Minneapolis, Minnesota, June 2019. Association for Computational Linguistics.
\newblock \doi{10.18653/v1/N19-4013}.
\newblock URL \url{https://aclanthology.org/N19-4013}.

\bibitem[Yu et~al.(2021)Yu, Xiong, and Callan]{yu2021improving}
HongChien Yu, Chenyan Xiong, and Jamie Callan.
\newblock Improving query representations for dense retrieval with pseudo relevance feedback.
\newblock In \emph{Proceedings of the 30th ACM International Conference on Information \& Knowledge Management}, pp.\  3592--3596, 2021.
\newblock URL \url{https://arxiv.org/abs/2112.06400}.

\bibitem[Yu et~al.(2023)Yu, Iter, Wang, Xu, Ju, Sanyal, Zhu, Zeng, and Jiang]{yu2023generate}
Wenhao Yu, Dan Iter, Shuohang Wang, Yichong Xu, Mingxuan Ju, Soumya Sanyal, Chenguang Zhu, Michael Zeng, and Meng Jiang.
\newblock Generate rather than retrieve: Large language models are strong context generators.
\newblock In \emph{The Eleventh International Conference on Learning Representations}, 2023.

\bibitem[Zhang et~al.(2021)Zhang, Ma, Shi, and Lin]{mrtydi}
Xinyu Zhang, Xueguang Ma, Peng Shi, and Jimmy Lin.
\newblock {Mr. TyDi}: A multi-lingual benchmark for dense retrieval.
\newblock \emph{arXiv:2108.08787}, 2021.
\newblock URL \url{https://arxiv.org/abs/2108.08787}.

\bibitem[Zhao et~al.(2021)Zhao, Lu, and Lee]{sparta}
Tiancheng Zhao, Xiaopeng Lu, and Kyusong Lee.
\newblock {SPARTA}: Efficient open-domain question answering via sparse transformer matching retrieval.
\newblock In \emph{Proceedings of the 2021 Conference of the North American Chapter of the Association for Computational Linguistics: Human Language Technologies}, pp.\  565--575, Online, June 2021. Association for Computational Linguistics.
\newblock \doi{10.18653/v1/2021.naacl-main.47}.
\newblock URL \url{https://aclanthology.org/2021.naacl-main.47}.

\bibitem[Zhou et~al.(2022)Zhou, Shen, Geng, Tao, Xu, Long, Jiao, and Jiang]{zhou2022towards}
Yucheng Zhou, Tao Shen, Xiubo Geng, Chongyang Tao, Can Xu, Guodong Long, Binxing Jiao, and Daxin Jiang.
\newblock Towards robust ranker for text retrieval.
\newblock \emph{arXiv e-prints}, pp.\  arXiv--2206, 2022.
\newblock URL \url{https://arxiv.org/abs/2206.08063}.

\bibitem[Zhu et~al.(2021)Zhu, Lei, Wang, Zheng, Poria, and Chua]{zhu21qasurvey}
Fengbin Zhu, Wenqiang Lei, Chao Wang, Jianming Zheng, Soujanya Poria, and Tat{-}Seng Chua.
\newblock Retrieving and reading: {A} comprehensive survey on open-domain question answering.
\newblock \emph{CoRR}, abs/2101.00774, 2021.
\newblock URL \url{https://arxiv.org/abs/2101.00774}.

\bibitem[Zhuo(2023)]{zhuo2023large}
Terry~Yue Zhuo.
\newblock Large language models are state-of-the-art evaluators of code generation, 2023.
\newblock URL \url{https://arxiv.org/abs/2304.14317}.

\end{thebibliography}
